\documentclass[11pt]{article}

\usepackage[preprint]{acl}

\usepackage{times}
\usepackage{latexsym}

\usepackage[T1]{fontenc}

\usepackage[utf8]{inputenc}

\usepackage{microtype}

\usepackage{inconsolata}

\usepackage{graphicx}
\usepackage{booktabs}
\usepackage{longtable}
\usepackage{xcolor}
\usepackage{array}
\usepackage{multirow}
\usepackage{amsmath}
\usepackage{hyperref}
\usepackage{longtable}
\usepackage{enumitem}
\usepackage{seqsplit}
\usepackage[most]{tcolorbox}
\usepackage{lipsum}
\usepackage{geometry}
\usepackage{setspace}
\usepackage{tabularx}

\title{HAD: HAllucination Detection Language Models Based on a Comprehensive Hallucination Taxonomy}

\author {
    Fan Xu\textsuperscript{1},
    Xinyu Hu\textsuperscript{1},
    Zhenghan Yu\textsuperscript{1},
    Li Lin\textsuperscript{1},
    Xu Zhang\textsuperscript{1},\\
    \textbf{Yang Zhang\textsuperscript{2},
    Wei Zhou\textsuperscript{2},
    Jinjie Gu\textsuperscript{3},
    Xiaojun Wan\textsuperscript{1}}
\\
    \textsuperscript{1}Wangxuan Institute of Computer Technology, Peking University\\
    \textsuperscript{2}Alibaba Group
    \textsuperscript{3}Fudan University
\\
\small{\texttt{\{xufan2000,huxinyu,zhangxu,wanxiaojun\}@pku.edu.cn,
\{efsotr\_l,zhenghanyu\}@stu.pku.edu.cn}}\\
\small{\texttt{\{yaoling.zy,jinjie.gujj\}@antgroup.com, zhouwei546138922@126.com}}
}

\begin{document}
\maketitle
\begin{abstract}
The increasing reliance on natural language generation (NLG) models, particularly large language models, has raised concerns about the reliability and accuracy of their outputs. A key challenge is hallucination, where models produce plausible but incorrect information. As a result, hallucination detection has become a critical task.
In this work, we introduce a comprehensive hallucination taxonomy with 11 categories across various NLG tasks and propose the \textbf{HA}llucination \textbf{D}etection (\textbf{HAD}) models\footnote{\url{https://github.com/pku0xff/HAD}}, which integrate hallucination detection, span-level identification, and correction into a single inference process.
Trained on an elaborate synthetic dataset of about 90K samples, our \textbf{HAD} models are versatile and can be applied to various NLG tasks. We also carefully annotate a test set for hallucination detection, called \textbf{HADTest}, which contains 2,248 samples.
Evaluations on in-domain and out-of-domain test sets show that our HAD models generally outperform the existing baselines, achieving state-of-the-art results on HaluEval, FactCHD, and FaithBench, confirming their robustness and versatility.
\end{abstract}

\section{Introduction}
The rapid advancement of natural language generation (NLG) has been largely driven by the development of large language models (LLMs)~\cite{li2024pre}. These models, characterized by their ability to process and generate human-like text, have found applications across diverse domains~\cite{chkirbene2024large}, including content creation, customer support, and personalized education. Despite their impressive performance, LLMs still have limitations. One of the most concerning issues is the phenomenon known as \textbf{hallucination}, in which models produce outputs that seem credible but are factually incorrect or unfaithful to the provided context~\cite{huang2023survey}. This problem poses significant challenges for users relying on LLMs for accurate information, raising critical concerns about the trustworthiness and accountability of AI-generated content. As LLMs continue to evolve and integrate into various applications, addressing hallucination is paramount to ensuring the reliability of these technologies in real-world scenarios.

\begin{figure}[ht]
    \centering
    \includegraphics[width=1\linewidth]{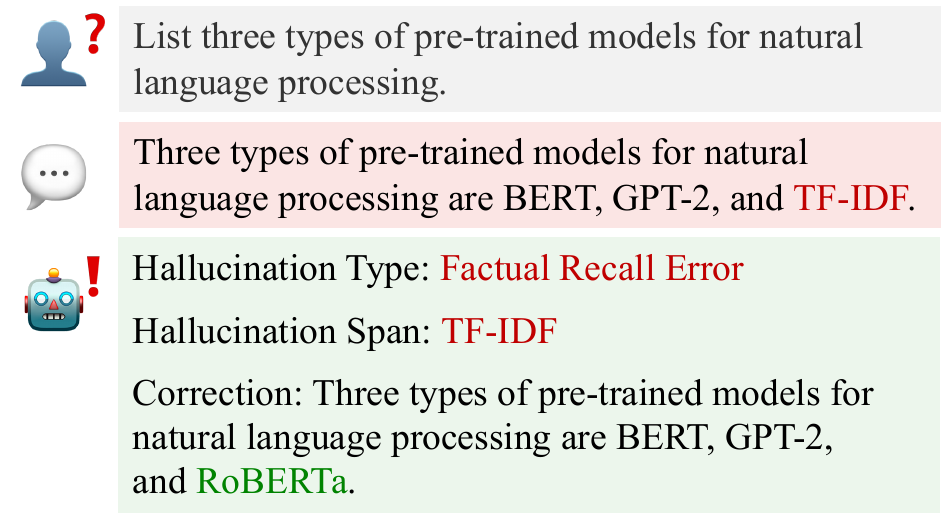}
    \caption{Example of the hallucination detection process.}
    \label{fig:demo}
\end{figure}

Hallucination in NLG can be broadly categorized into two primary dimensions: factuality and faithfulness \cite{huang2023survey}. \textbf{Factuality} refers to the degree to which the generated content aligns with real-world facts.
Meanwhile, \textbf{faithfulness} refers to the relationship between the generated text and the input, as well as the consistency within the output itself. It emphasizes the importance of following the instructions and accurately representing the information provided to the model.
Both aspects of hallucination impact the overall reliability of LLMs, so it's essential to understand and address both dimensions.

Despite the significant advancements in research on hallucination in LLMs, existing works still have notable limitations. Many studies focus solely on factuality~\cite{chen2024factchd} or faithfulness~\cite{niu2023ragtruth}, neglecting the comprehensive assessment of both aspects. Additionally, several prior studies target specific tasks~\cite{gu2024anah,min-etal-2023-factscore}, which restricts their applicability and generalizability. Furthermore, the taxonomy they employed is typically coarse-grained, lacking the nuanced granularity necessary for detailed analysis and effective application across diverse contexts.

In this work, we propose a more comprehensive and fine-grained hallucination taxonomy (Figure~\ref{fig:taxonomy}), covering both faithful and factual aspects. 
Our taxonomy is organized into three hierarchical levels and defines 11 specific types of hallucinations. Recognizing that NLG tasks vary in their requirements for these dimensions~\cite{ji2023survey}, we map each task type to corresponding hallucination categories.
Based on this correspondence, we synthesize a set of 90,172 training data samples by modifying correct task data to include various types of hallucinations across multiple NLG tasks. The synthetic data consists of hallucinated outputs and the spans of hallucinations as its key components.

Our \textbf{HA}llucination \textbf{D}etection (\textbf{HAD}) models are obtained by fine-tuning \texttt{Qwen2.5-7B-Instruct} and \texttt{Qwen2.5-14B-Instruct}~\cite{qwen2.5} with synthetic data, enabling them to jointly perform type classification, span-level identification, and correction within a unified inference.
To obtain high-quality test data, we manually annotate the \textbf{HADTest} dataset, including diverse hallucination samples paired with an equal number of correct samples, with 2,248 samples in total. We evaluate our models and other existing baselines through both in-domain and out-of-domain tests. The results show that HAD models outperform both general-purpose base models and specified hallucination detection models. The in-domain results demonstrate their functionality, while the out-of-domain results highlight their accuracy and generalizability. The datasets and models will be publicly available soon.

Overall, our main contributions can be summarized as follows: 
\begin{enumerate}[noitemsep, topsep=0pt, leftmargin=*]
    \item We present a fine-grained hallucination taxonomy that encompasses both factuality and faithfulness, considering more subtle issues related to hallucinations.
    
    \item Aligned with our hallucination taxonomy, we create a large-scale, multi-task training dataset along with a high-quality test set, \textbf{HADTest} for the hallucination detection and correction.
    
    \item We develop the \textbf{HAD} models that are capable of handling hallucination categorization, span-level detection, and correction within a single inference. These models are applicable to various NLG tasks and have achieved state-of-the-art performance on multiple benchmarks.
\end{enumerate}

\section{Hallucination Taxonomy}

\begin{figure*}[ht]
    \centering
    \includegraphics[width=\linewidth]{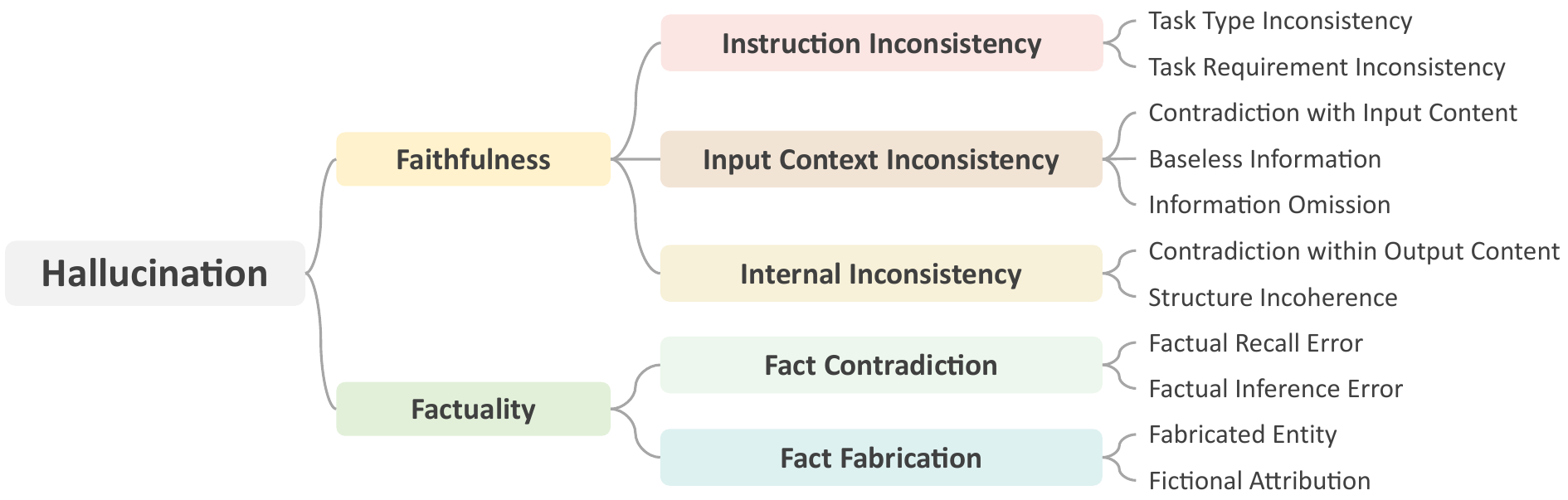}
    \caption{Taxonomy of hallucinations, comprising three levels and 11 fine-grained categories.}
    \label{fig:taxonomy}
\end{figure*}

We define \textbf{hallucination} as the phenomenon where the generated output appears credible but is factually incorrect or not faithful to the provided context, following the taxonomy proposed in previous work \cite{huang2023survey}.
Typically, A natural language generation task consists of the following components:
\begin{itemize}[noitemsep, topsep=0pt, leftmargin=*]
    \item \textbf{Instruction:} Description of the task, indicating the task type and specific constraints, such as the subject matter, format, and length of the generated output.
    \item \textbf{Input context:} The relevant context for generating the output, such as the source document in summarization, translation, or contextual QA.
    \item \textbf{Task Output:} The result generated to complete the task.
\end{itemize}

We categorize hallucination hierarchically based on the nature of the inconsistencies and the content involved, as shown in Figure \ref{fig:taxonomy}. This taxonomy includes 11 fine-grained categories at the third level. Compared to existing hallucination taxonomies, ours is more comprehensive, covering a wider range of scenarios and providing a more detailed analysis of hallucinations. Examples of different hallucination types are presented in Table \ref{tab:hallu_examples} in Appendix \ref{sec:supp_taxonomy}. We construct this taxonomy by synthesizing insights from a broad range of existing research on hallucinations in LLMs. To demonstrate the empirical relevance and coverage of this taxonomy, we map its categories to related concepts and terminologies discussed in prior works, as presented in Table \ref{tab:type_mapping} in Appendix~\ref{sec:supp_taxonomy}. Below, we present a detailed definition of each category.

\subsection{Faithfulness Hallucination}
Faithfulness hallucinations stem from inconsistencies between the input content and generated content, or from issues within the generated content, without relying on external factual information. These hallucinations can be classified into three main types: instruction inconsistency, input context inconsistency, and internal inconsistency. 

\paragraph{\textbf{A. Instruction Inconsistency}} This occurs when generated content fails to align with the given instruction, not meeting task, content, or format requirements.
\begin{enumerate}[noitemsep, topsep=0pt, leftmargin=*]
    \item \textbf{Task Type Inconsistency (TTI)}: The output represents a different task type than instructed. Deviations within the same task type, such as violating specific requirements, are excluded.

    \item \textbf{Task Requirement Inconsistency (TRI)}: The output doesn't meet specified task requirements, such as format, length, subject, or tone. This is due to a failure to follow instructions, rather than to contradictions in content.
\end{enumerate}

\paragraph{\textbf{B. Input Context Inconsistency}} This happens when the generated content contradicts or misinterprets the input context.

\begin{enumerate}[resume, noitemsep, topsep=0pt, leftmargin=*]
    \item \textbf{Contradiction with Input Content (CwIC)}: The output contradicts the input, presenting information incompatible with the context, often due to failure to recall or misunderstanding the provided information.

    \item \textbf{Baseless Information (BI)}: In tasks that require strict adherence to the given context, the output contains unsupported information. Tasks that seek new information do not suffer from this problem.

    \item \textbf{Information Omission (IO)}: When the task requires a complete and accurate representation of the provided context, the output omits details presented in the input.
\end{enumerate}

\paragraph{\textbf{C. Internal Inconsistency}} This arises when generated content contains contradictions, logical errors, or structural problems, leading to incoherent or implausible results.

\begin{enumerate}[resume, noitemsep, topsep=0pt, leftmargin=*]
    \item \textbf{Contradiction within Output Content (CwOC)}: The output includes contradictory statements or flawed reasoning.
    
    \item \textbf{Structural Incoherence (SI)}: The output contains redundant, repetitive, or disjointed statements that do not enhance clarity or value.
\end{enumerate}

\subsection{Factuality Hallucination}
Factuality hallucinations occur when the generated content contains inaccuracies, distortions, or fabrications that do not align with external reality. According to whether they can be directly refuted by established world knowledge, factuality hallucinations can be categorized into fact contradiction and fact fabrication.

\paragraph{\textbf{D. Fact Contradiction}} Fact contradiction happens when the generated content directly contradicts established knowledge.

\begin{enumerate}[resume, noitemsep, topsep=0pt, leftmargin=*]
    \item \textbf{Factual Recall Error (FRE)}: The generated text contains an incorrect atomic fact due to the model's inability to accurately recall or access relevant knowledge.

    \item \textbf{Factual Inference Error (FIE)}: The content contains incomplete or misinterpreted facts, such as confusion between time periods, individuals, or events, omission of key details, or errors in the order of causes and effects, leading to the facts being presented incorrectly.
\end{enumerate}

\paragraph{\textbf{E. Fact Fabrication}} Fact fabrication refers to content presenting unverifiable information not based on real-world knowledge, excluding artistic or creative fiction.

\begin{enumerate}[resume, noitemsep, topsep=0pt, leftmargin=*, align=left]
    \item \textbf{Fabricated Entity (FE)}: The generated output introduces entirely new, fabricated entities such as concepts, names, or objects that lack any real-world basis. 
    \item \textbf{Fictional Attribution (FA)}: The generated output fabricates information about real entities, such as unverified claims or quotes, that cannot be directly confirmed or disproven by reliable sources. Unlike a Fabricated Entity, this error does not introduce entirely new entities.
\end{enumerate}

\section{Data Construction}
\label{subsec:task_formulation}
The goal of our HAD models is to provide a fine-grained detection and classification of hallucination, and then correct it. Given a \textbf{task input} and a \textbf{task output}, the model is required to classify the hallucination into a fine-grained \textbf{type}, identify the precise \textbf{span} of the hallucinated content, and provide a \textbf{correction} to the output. The task input is a combination of instruction and input context, considering that in many scenarios, people do not explicitly distinguish them. Our models aim to apply to multiple NLG tasks and hallucination types, which places high demands on data diversity. In the following subsections, we will introduce how we acquire the training and test data.

\subsection{Source Data Selection}
Different tasks can lead to various types of hallucinations~\cite{ji2023survey}, so we categorize NLG tasks into the following four types.
\begin{itemize}[noitemsep, topsep=0pt, leftmargin=*]
    \item \textbf{Information Expansion}: These tasks involve generating output that adds significant details to the input, often producing longer text. They require maintaining coherence and logical consistency, as seen in story writing. Knowledge-seeking tasks, such as long-form question answering (LFQA), require factual accuracy.
    \item \textbf{Information Alignment}: The goal is to reflect the input information accurately without adding, omitting or altering any details. Tasks like paraphrasing and data-to-text generation fall under this category.
    \item \textbf{Information Condensation}: The task is to condense or extract key details from the input, ensuring conciseness while retaining core meaning. Summarization and contextual QA are examples of this type.
    \item \textbf{Information Continuation}: This involves continuing the input information coherently, avoiding contradictions or inconsistencies. Short-form QA, math reasoning, dialogue, and instruction following are typical tasks in this category. The requirements for factual accuracy vary depending on the domain.
\end{itemize}

We adopt ELI5~\cite{fan-etal-2019-eli5} as the data source for the LFQA task. From the Super-NaturalInstruction (SNI) dataset \cite{wang-etal-2022-super}, we sample data for tasks including story writing, poem writing, paraphrasing, data-to-text, summarization, contextual QA and short-form QA. For math reasoning, the data source is GSM8K \cite{cobbe2021gsm8k}. Data of dialogue task comes from the FaithDial dataset \cite{dziri-etal-2022-faithdial}. Additionally, we use Alpaca~\cite{alpaca} for the general instruction following task.

\subsection{Hallucination Synthesis}
\label{sec:hallucination_synthesis}
\paragraph{Hallucination Injection}
We synthesize the hallucination data from multiple sources introduced above.
For each dataset, we manually assess which hallucination types may occur based on the task definition and the task categories mentioned above. All hallucination data are constructed with GPT-4o(\texttt{gpt-4o-2024-08-06}) by disturbing the correct output based on the type definition and 3-shot examples. The examples differ by the hallucination type. To improve the variety, the examples for each data item are sampled independently. The prompt template can be found in Table \ref{tab:injection_prompt} in Appendix \ref{sec:prompt_template}. We sample 10,000 source data for each hallucination type, set the temperature to 1.0, and sample 5 hallucination candidates for each data item.

\paragraph{Automatic Filtering}
\label{para:filering}
After the hallucination injection, we check and filter the candidates with a few criteria, which vary between hallucination types.
Some issues may exist in the injection stage, including the failure to follow instructions, misunderstanding the hallucination type, or misidentifying the hallucination span.
To address these issues, we write both general and task-specific criteria and prompt GPT-4o to analyze whether all the criteria are met. The prompt template and criterion for data verification are shown in Table \ref{tab:check_prompt} and \ref{tab:check_criterion} in Appendix \ref{sec:prompt_template}.
The pass rate at the filtering stage is 82.3\%. After filtering, there are 90088 hallucination data left.

\subsection{Test Data Annotation}
\label{sec:test_set}
\begin{figure}
    \centering
    \includegraphics[width=1.0\linewidth]{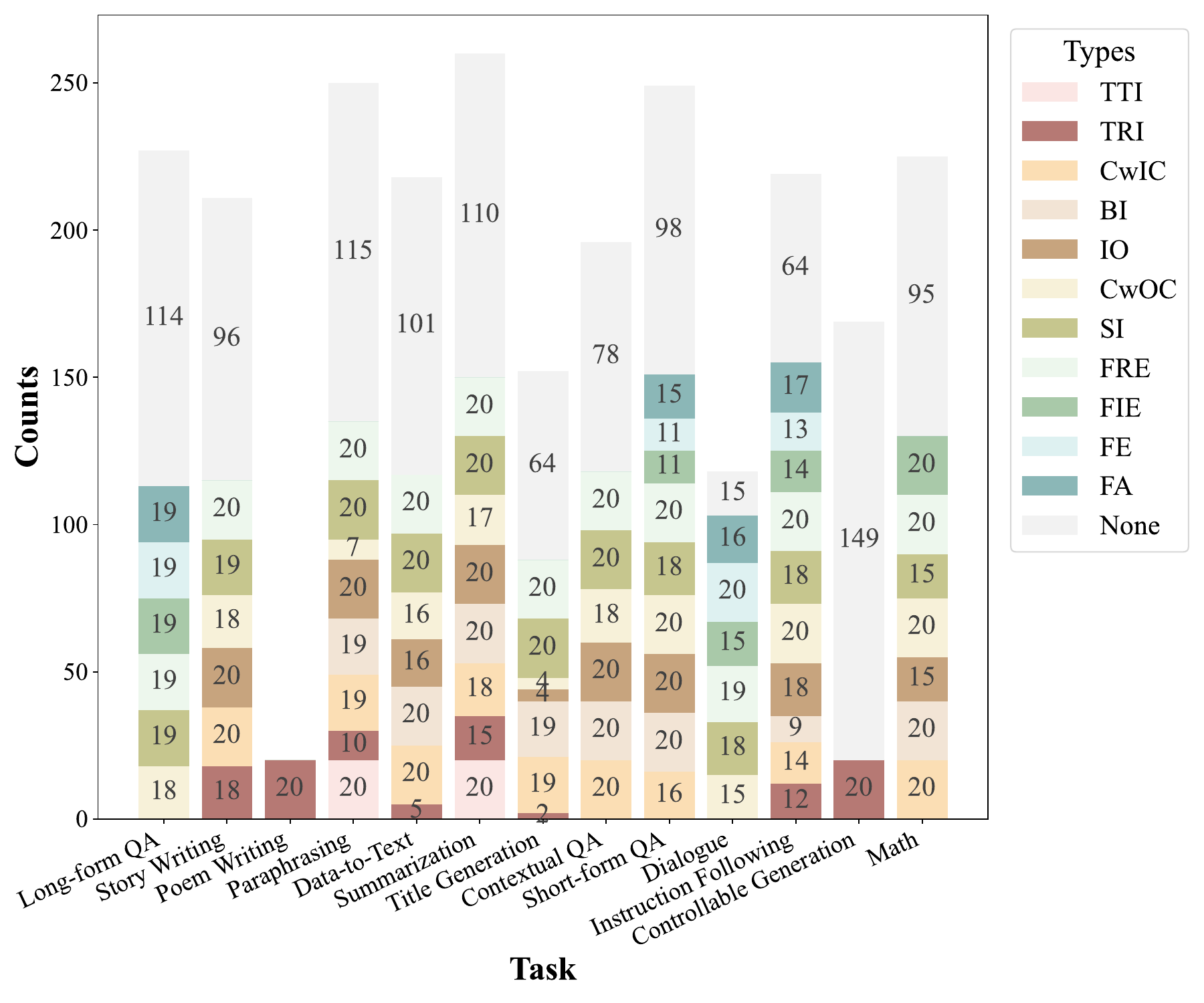}
    \caption{Statistics of our constructed HADTest.}
    \label{fig:test_statistics}
\end{figure}

We annotate a test set of 2,248 samples, referred to as \textbf{HADTest}. To construct this set, we first sample 20 data items for each task and hallucination type from the previously constructed data, resulting in a raw test dataset of 1,240 samples. To ensure high quality, the annotation was conducted by the authors themselves, who possess substantial expertise in large language models and specifically in the study of hallucination. Each data item was independently assessed by two annotators to determine whether the injected hallucination conformed to the predefined type definition and simultaneously satisfied all additional criteria specified in Table \ref{tab:check_criterion} (Appendix \ref{sec:prompt_template}). The criteria also help distinguish between easily confusable hallucination types.
A data item was retained only when both annotators reached agreement. For cases that did not fully meet the requirements, we carefully edited the hallucinated output and span to generate qualified hallucination data whenever possible. The pass rate for the raw test data is 66.37\%, and the inter-annotator agreement rate is 80.56\%. After filtering and editing, the final hallucinated portion of HADTest consists of 1,124 samples. We then add an equal number of positive (non-hallucinated) data samples, resulting in a balanced test set of 2,248 samples in total. Detailed statistics are shown in Figure \ref{fig:test_statistics}.

\section{Experiments}
\subsection{Settings}
We adopt end-to-end supervised fine-tuning to enable the models to jointly perform hallucination categorization, span-level detection, and correction. The prompt template and response template are provided in Table \ref{tab:detection_prompt} in Appendix \ref{sec:prompt_template}, based on the task formulation outlined in Section \ref{subsec:task_formulation}.
The hallucination types and spans are synthesized following the procedure described in Section \ref{sec:hallucination_synthesis}, while the correction is derived from the original ground-truth output of the source datasets without further processing.
We sampled 90,172 data points for training and 869 for development.
The data ratio in the hallucination dataset reflects both the frequency of hallucination types in real scenarios and their detection difficulty.
Specifically, there are about 9,000 samples for each factual hallucination type, around 5,000 for input context inconsistencies, and roughly 2,500 for other types.
Positive data are sampled at half the amount of hallucination data while maintaining consistent task distributions.

Our primary base model is \texttt{Qwen-2.5-14B-} \texttt{Instruct}. We also experiment with \texttt{Qwen-2.5-} \texttt{7B-Instruct} and \texttt{Llama-3.1-8B-Instruct} to assess performance across different model sizes and architectures. We perform supervised fine-tuning on these models using the training data for 1 epoch with a learning rate of \texttt{1e-5} and a batch size of \texttt{256}. The devices used are 4 H100 GPUs. We set the temperature to 0 at the test stage.

In addition to the original HAD models, we introduce a binary classification version called HAD-14B-Binary. The main difference is that it simplifies the fine-grained classification into a binary decision: whether the text contains a hallucination or not. HAD-14B-Binary is trained with the same data composition and parameter settings as the HAD-14B model.
The prompt and response templates are provided in Table \ref{tab:binary_detection_prompt} in Appendix \ref{sec:prompt_template}.

\subsection{Evaluation}
\paragraph{In-Domain Evaluation} We use HADTest for in-domain testing and evaluate three main aspects: classification, span detection, and correction. 
For the evaluation of hallucination type classification, we report \textbf{Accuracy}, \textbf{Balanced Accuracy (BA)}, and \textbf{Macro-F1} on the fine-grained labels. To provide a binary perspective, we further collapse the fine-grained categories into two classes (hallucination vs.\ non-hallucination) and report \textbf{Accuracy} on this binary task. For span detection and correction, we assess performance using word-level \textbf{Precision}, \textbf{Recall}, and \textbf{F1 score}, and present the average results across predictions. Both span detection and correction are evaluated exclusively on samples containing hallucinations.

\paragraph{Out-of-Domain Evaluation} For the out-of-domain (OOD) test, we select the following datasets and benchmarks:
\begin{itemize}[noitemsep, topsep=0pt, leftmargin=*]
    \item \textbf{HaluEval}~\citep{li-etal-2023-halueval}, a large-scale hallucination benchmark that covers four tasks: instruction following, dialogue, question answering, and summarization.
    \item \textbf{FactCHD}~\citep{chen2024factchd}, a QA dataset focused on fact-conflicting hallucinations with diverse patterns.
    \item \textbf{FaithBench}~\citep{bao2024faithbench}, a summarization benchmark for faithfulness hallucination.
\end{itemize}

Here we simplify the hallucination detection to a binary classification task, i.e., whether the task output contains hallucination.
The metrics used are \textbf{Accuracy} for HaluEval, \textbf{Micro F1} for positive samples in FactCHD, and both \textbf{Balanced Accuracy} and \textbf{Macro F1} for FaithBench.

\paragraph{Baselines}
We select leading base LLMs as part of the baselines.
The selected models are \textbf{GPT-4o}\footnote{https://openai.com/index/gpt-4o-system-card/}, \textbf{GPT-4o mini}\footnote{https://openai.com/index/gpt-4o-mini-advancing-cost-efficient-intelligence/} and \textbf{DeepSeek V3}~\cite{deepseekai2024deepseekv3technicalreport}.

We also select several hallucination detection models and methods as baselines:
\begin{itemize}[noitemsep, topsep=0pt, leftmargin=*]
    \item \textbf{SelfCheckGPT}~\cite{manakul-etal-2023-selfcheckgpt} is a representative sampling-based method for detecting hallucinations and can be applied to black-box models without requiring external references.
    \item \textbf{LYNX 8B}\footnote{https://huggingface.co/PatronusAI/Llama-3-Patronus-Lynx-8B-Instruct-v1.1}~\cite{ravi2024lynx}, a hallucination detection model designed for retrieval-augmented generation scenarios. It only supports binary classification.
    \item \textbf{ANAH-v2}~\cite{gu2024anah}, a model tailored for hallucination annotation in question-answering tasks. The output label is \textit{No Hallucination}, \textit{Contradictory Hallucination}, \textit{Unverifiable Hallucination}, or \textit{No Fact}, and we treat the last two labels as 0.5 when calculating metrics for binary classification.
    \item \textbf{FAVA-Model}~\cite{mishra2024fine}, a model that detects and corrects fine-grained hallucinations.
    \item \textbf{HHEM-2.1-Open}\footnote{https://huggingface.co/vectara/hallucination\_evaluation\\\_model}~\cite{hhem-2.1-open}, a hallucination evaluation model that generates a consistency score between two pieces of text.
\end{itemize}

More details about the implementation of baselines can be found in Appendix \ref{sec:implementation_of_baselines}.

\subsection{Results}
\begin{table*}[ht]
    \centering
    \small
    \setlength\tabcolsep{6.3pt}
    \begin{tabular}{l|c|ccc|ccc|ccc}
        \toprule
        \multirow{2}*{\textbf{Model}} & \textbf{Binary} & \multicolumn{3}{c|}{\textbf{Fine-Grained}} & \multicolumn{3}{c|}{\textbf{Span}} &\multicolumn{3}{c}{\textbf{Correction}}\\
         & \textbf{Acc} & \textbf{Acc} & \textbf{BA} & \textbf{F1} & \textbf{Precision} & \textbf{Recall} & \textbf{F1} & \textbf{Precision} & \textbf{Recall} & \textbf{F1} \\
         \midrule
        FAVA-Model & 59.43 & - & - & - & 34.69 & 50.35 & 36.56 & 70.84 & 65.41 & 66.29 \\
        \midrule
        GPT-4o & 63.08 & 40.30 & 46.61 & 41.64 & 56.41 & 73.03 & 57.79 & 63.22 & 68.93 & 62.90 \\
        GPT-4o mini & 60.19 & 29.76 & 32.00 & 29.37 & 48.00 & 59.03 & 46.91 & 57.99 & 65.17 & 58.81 \\
        DeepSeek-V3 & 58.50 & 39.72 & 38.48 & 38.47 & 46.58 & 60.45 & 48.15 & 40.41 & 63.23 & 45.37 \\
        \midrule
        HAD-7B & 87.46 & 81.76 & 76.09 & 75.17 & 76.50 & 74.08 & 72.52 & 74.90 & 74.41 & 74.05\\
        HAD-8B & 86.12 & 79.72 & \textbf{77.74} & 74.09 & 78.73 & 77.07 & 75.21 & 79.03 & 78.60 & 78.29 \\
        HAD-14B & \textbf{89.10} & \textbf{83.05} & 77.38 & \textbf{76.29} & 78.96 & 78.27 & 76.01 & 78.87 & 78.16 & 77.97 \\
        HAD-14B-Binary & 87.77 & - & - & - & \textbf{82.48} & \textbf{80.95} & \textbf{78.73} & \textbf{82.80} & \textbf{82.01} & \textbf{81.88} \\
        \bottomrule
    \end{tabular}
    \caption{The in-domain test results on the HADTest dataset include evaluations for binary classification, fine-grained classification, span identification, and correction. A dash ("–") indicates that the model does not support the corresponding function.}
    \label{tab:iid_results}
\end{table*}

\begin{table*}[ht]
\centering
\small
\setlength\tabcolsep{4pt}
\begin{tabular}{l|c|c|c|c|c|cc}
\toprule
\multirow{2}*{\textbf{Model}} & \textbf{HaluEval-dial} & \textbf{HaluEval-gen} & \textbf{HaluEval-QA} & \textbf{HaluEval-summ} & \textbf{FactCHD} & \multicolumn{2}{c}{\textbf{FaithBench}} \\
& \textbf{Acc} & \textbf{Acc} & \textbf{Acc} & \textbf{Acc} & \textbf{Micro F1} & \textbf{BA} & \textbf{Macro F1} \\

\midrule
SelfCheckGPT & 48.20 & 21.30 & 36.30 & 49.70 & 54.54 & 50.00 & 41.26 \\
LYNX 8B & 60.25 & 74.48 & 85.70* & 68.45 & 44.88 & 56.71 & 44.36 \\
ANAH-v2 & \textcolor{gray}{57.00} & \textcolor{gray}{59.98} & \textcolor{gray}{81.21*} & \textcolor{gray}{57.33} & \textcolor{gray}{57.44} &\textcolor{gray}{51.03} &\textcolor{gray}{26.68} \\
FAVA-Model & 57.82 & 73.37 & 62.16 & 50.42 & 44.62 & 53.18 & 41.64 \\
HHEM-2.1-Open & - & - & - & 37.86 & - & 55.68* & 40.86* \\
\midrule
GPT-4o & 73.55 & 81.30 & 86.33 & 71.43 & 51.89 & 56.29* & 40.75* \\
GPT-4o mini & 71.71 & 81.18 & 84.04 & 69.80 & 39.20 & 52.13 & 35.29 \\
DeepSeek-V3 & 77.85 & \textbf{81.81} & 50.73 & 72.09 & 62.98 & 52.94 & 32.79 \\
\midrule
HAD-7B & 82.24 & 78.41 & 83.63 & 81.54 & 63.68 & 44.63 & 44.30 \\
HAD-8B & \textbf{84.55} & 75.40 & 81.40 & 76.55 & 64.16 & 55.46 & 55.55 \\ 
HAD-14B & 82.33 & 79.94 & 86.65 & 81.36 & \textbf{66.82} & 51.85 & 45.45 \\
HAD-14B-Binary & 82.16 & 76.55 & \textbf{92.37} & \textbf{84.63} & 60.18 & \textbf{57.74} & \textbf{54.97} \\
\bottomrule
\end{tabular}
\caption{Model performance on the out-of-domain test sets. Note that the results with '*' is copied from the original papers. In the ANAH-v2 model test, we classify samples with the model output of "unverifiable" or "nofact" as 0.5 non-hallucinated and 0.5 hallucinated samples (highlighted in gray).}
\label{tab:ood_results}
\end{table*}

Overall, HAD-14B and HAD-14B-Binary outperform the baseline models across multiple evaluation metrics and test sets.
Table \ref{tab:iid_results} presents the in-domain test results. HAD-14B delivers superior performance across multiple metrics, achieving a binary classification accuracy of 89.10\% and a fine-grained classification accuracy of 83.05\%. HAD-14B also excels in span identification, with an F1 score of 76.01\%, and in correction, with an F1 score of 77.97\%. More results can be found in Appendix \ref{sec:supp_results}.

As seen in Table \ref{tab:ood_results}, HAD models outperform or are comparable with both closed-source, large-scale baseline models and other hallucination detection methods across most OOD test sets, highlighting their robustness. However, its performance on the HaluEval-general datasets is lower, suggesting the need for further investigation into the model's limitations.
HAD-7B, despite its smaller size, still outperforms baseline models on certain test sets.
There is also a trade-off between detection accuracy and functionality, as the performance of HAD-14B-Binary outperforms that of HAD-14B on several test sets.

\subsection{Error Analysis}
\begin{figure}[ht]
    \centering
    \includegraphics[width=\linewidth]{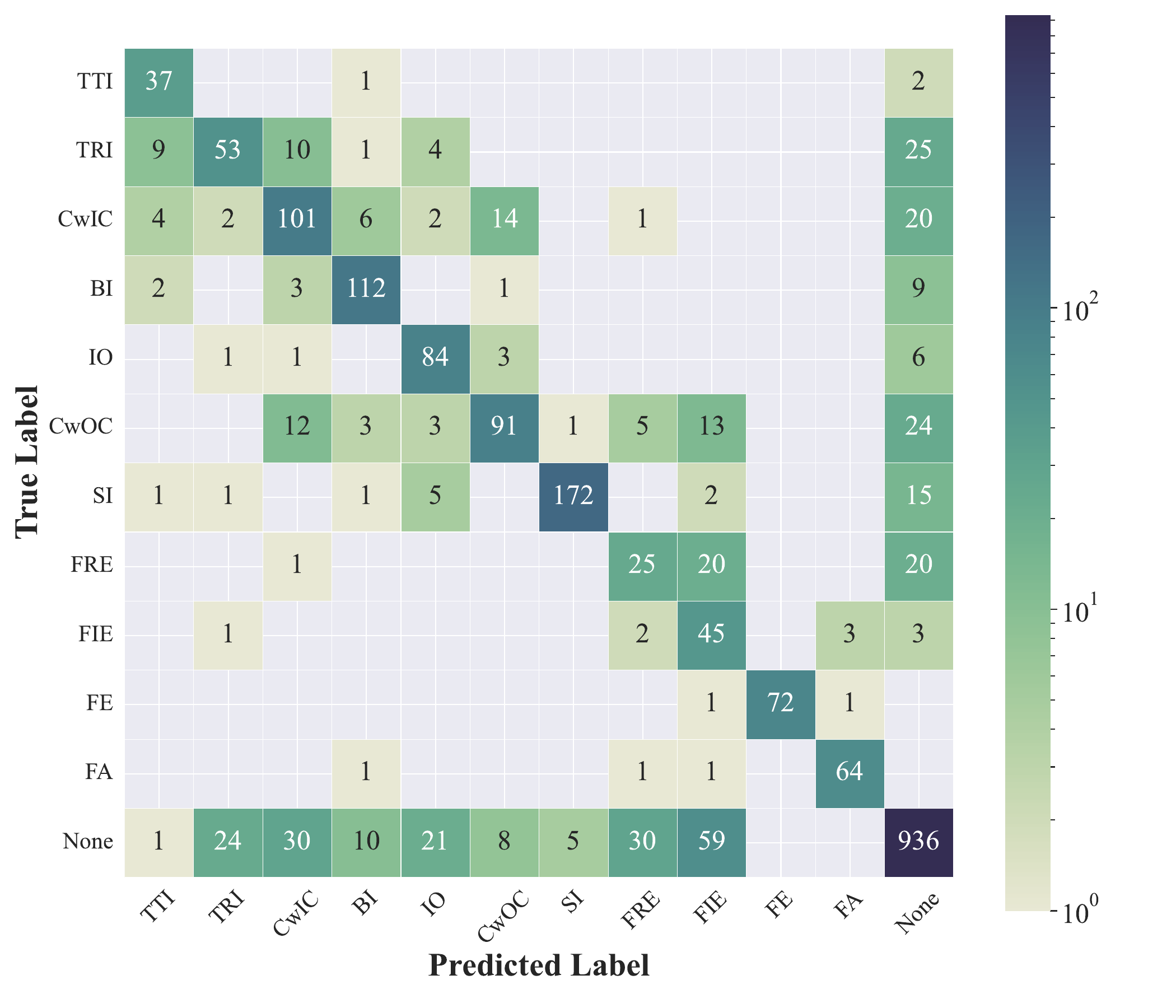}
    \caption{Confusion matrix of the test results generated by HAD-14B.}
    \label{fig:confusion_matrix}
\end{figure}

We use the results of HAD-14B for this section and analyze the in-domain test results across different types. 
The confusion matrix is shown in Figure \ref{fig:confusion_matrix}.
The F1 scores of \textit{Factual Recall Error} and \textit{Factual Inference Error} are relatively low (35.90 and 26.87, respectively) because these errors are often mistakenly labeled as "no hallucination." These failures are mainly due to insufficient background knowledge, particularly in smaller models.
Besides, there’s noticeable confusion between similar categories, such as \textit{Contradiction with Input Content} and \textit{Contradiction within Output Content}, as well as between \textit{Factual Inference Error} and \textit{Factual Recall Error}.

In the HaluEval-general dataset, false negatives are mainly classified as factual recall errors or task requirement inconsistencies. Additionally, this dataset has a wider range of task types and requirements and contains responses generated by real LLMs, which makes hallucination detection more challenging, resulting in more false positives.

In summary, the error analysis highlights that HAD-14B still struggles with fact contradiction hallucinations, which are largely attributed to a lack of background knowledge, especially in smaller models. The analysis also suggests areas for improvement, such as better handling of diverse tasks and enhancing the model's ability to distinguish between different types of hallucinations.

\subsection{Ablation Study}
The hallucination verification step in Section \ref{para:filering} aims to improve data quality. To evaluate its impact, we randomly sample an equal number of data points with the same distribution from the raw hallucination data and train a separate model for comparison. We evaluate it with the fine-grained classification task and the macro F1 on our test set is 75.65\%, which is lower than the HAD-14B's accuracy of 76.29\%.

\section{Knowledge Augmentation}
Due to the limitations of the language model itself, including the cutoff time of its training data and its limited knowledge base, background knowledge retrieval has become an important auxiliary tool for detecting hallucinations, particularly in identifying factuality hallucinations. Thus, we integrate retrieved information into the hallucination detection process by simply inserting background knowledge into the \texttt{task input} part. We concatenate the task input and task output with a "\texttt{\textbackslash n}" as the query and use the \texttt{contriever-msmarco} model~\cite{izacard2021contriever} to retrieve relevant paragraph from the Wiki database.

\begin{table}[ht]
    \centering
    \small
    \setlength\tabcolsep{9pt}
    \begin{tabular}{l|cccc}
        \toprule
        \textbf{Method} & \textbf{FRE} & \textbf{FIE} & \textbf{FE} & \textbf{FA} \\
        \midrule
        HAD-14B & 35.90 & 26.87 & 81.48 & 70.48 \\
        +knowledge & 37.84 & 52.17 & 97.99 & 87.88 \\
        \bottomrule
    \end{tabular}
    \caption{F1-score of in-domain test results with knowledge augmentation.}
    \label{tab:iid_ret_results}
\end{table}

We test knowledge augmentation for HAD-14B and report the F1 score of four fact hallucination types for the HADTest dataset in Table \ref{tab:iid_ret_results}.
As expected, knowledge augmentation can largely improve the model performance on fact-checking, on both in-domain and out-of-domain test sets.
The improvement in the Factual Inference Error is the most obvious, indicating that the model is particularly dependent on external knowledge when dealing with the association of multiple facts.

Given that HAD models are built on 7B and 14B base models with limited knowledge, augmenting them with external information is crucial and effective for fact-based domains. Our experiments demonstrate a simple yet powerful approach to implementing this.

\section{Related Work}
\subsection{Hallucination Detection Model}
Training a language model to detect hallucinations is a cost-effective approach to some extent. Several recent works have made significant strides in this area by proposing hallucination detection models and benchmarks.
For instance, FAVABench~\cite{mishra2024fine} is a dataset designed to evaluate hallucinations in large language models. This work classifies six types of hallucinations and proposes FAVA, a retrieval-augmented model trained on Llama2-Chat 7B to detect and edit hallucinations. It is the closest to ours, but is limited to factual hallucinations with mandatory retrieval and focuses solely on fact-seeking tasks. Several other works, including RAGTruth~\cite{niu2023ragtruth}, FactCHD~\cite{chen2024factchd} and ANAH-v2~\cite{gu2024anah} adopt the same workflow that propose a hallucination dataset and then train an accompanying hallucination detection model.
LYNX~\cite{ravi2024lynx}, a series of models, is designed for large-scale hallucination detection in contextual question-answering tasks within an RAG framework.
These models, while performing excellently on specific tasks or domains, face significant challenges when dealing with diverse data from various tasks or types of hallucinations. In contrast, our HAD is designed to handle a broader range of tasks and hallucination types, offering greater versatility and applicability across a variety of use cases.

\subsection{Taxonomy of Hallucination}
To better understand and address hallucination in natural language generation, establishing a detailed taxonomy is essential. Several surveys \cite{ji2023survey,ye2023cognitive} divide hallucinations into two parts: intrinsic hallucinations, where the output contradicts the source content, and extrinsic hallucinations, where the output cannot be verified by the source content. In another survey~\cite{zhang2023siren}, hallucinations are categorized into three types based on where the outputs conflict: input-conflicting, context-conflicting, and fact-conflicting. Though these taxonomies are valuable for further investigations, they lack a fine-grained taxonomy of hallucinations.
We inherit and further expand the taxonomy from \cite{huang2023survey}, providing a more detailed analysis of the hallucination problem.

\subsection{Hallucination Benchmark}
As hallucination has become a significant area of investigation into large language models (LLMs), numerous benchmarks have been developed.
For example, HalluQA \cite{cheng2023evaluating} contains three types of questions (misleading, misleading-hard, and knowledge), aiming to detect hallucination generation in Chinese LLMs. 
Moreover, some benchmarks are developed to assess the ability of LLMs to detect hallucinations. HaluEval \cite{li-etal-2023-halueval} consists of 5,000 user queries and 30,000 task-specific samples generated from LLMs' automatic generation and human annotation. FELM \cite{zhao2024felm} can be used to assess LLMs' factuality in five domains, including world knowledge, science and technology, mathematics, writing and recommendation, and reasoning. 
Although these studies are compelling, there is still a need for a large-scale, multi-task hallucination dataset for training. In our study, we aim to provide a dataset covering multiple NLG tasks and hallucination categories.

\section{Conclusion}
In conclusion, we propose a fine-grained hallucination taxonomy with 11 fine-grained categories and construct a large dataset for training hallucination detection models covering multiple NLG tasks. Additionally, we manually annotate a test set called HADTest. Based on the hallucination taxonomy and data, we introduce HAD models for hallucination detection. Beyond simply detecting whether there are hallucinations in the text, HAD models can also classify fine-grained hallucination types, locate their spans, and correct them. Our models outperform existing general-purpose models as well as hallucination detection models, offering improved functionality, detection accuracy, and applicability across different NLG tasks.

\section*{Limitations}
Although our proposed HAD models achieve good performance on both in-domain and out-of-domain benchmarks, several limitations remain. First, the current formulation is restricted to detecting single-span hallucinations, whereas real-world outputs often contain multiple or overlapping erroneous spans, as well as different hallucination types within the same task output. A possible remedy is to first decompose outputs into smaller units before applying the HAD model. 

Second, the training pipeline depends heavily on synthetic hallucination data, which, despite its scalability, cannot fully capture the diversity and complexity of naturally occurring cases. However, annotating natural hallucinations is costly and difficult to control in terms of distribution, making this trade-off between synthetic and natural data persistent and difficult to resolve. 

On the methodological level, the functionality of our model is primarily driven by the composition and diversity of the data. We have not fully explored how to coordinate between multiple tasks and various types of hallucinations to achieve optimal results.

\bibliography{detection_custom}

\appendix
\label{sec:appendix}
\section{Supplementary Materials for Hallucination Taxonomy}
\label{sec:supp_taxonomy}
We build our hallucination taxonomy based on prior works about LLMs, as shown in Table \ref{tab:type_mapping}, each category can be find similar concepts in related research. This mapping provide important supporting for the coverage and rationality of our taxonomy.
To help better understand the meaning of each hallucination type, we provide examples in Table \ref{tab:hallu_examples}.

\section{Implementation of Baselines}
\label{sec:implementation_of_baselines}
When testing base models on our in-domain HADTest dataset, we provide hallucination-type descriptions and fixed 2-shot examples in the prompt (Table \ref{tab:baseline_detection_prompt}). One example is correct and another is hallucinated. For out-of-domain test datasets, we adopt the same prompts as specified in the original papers corresponding to these datasets, with the temperature parameter set to 1.

For SelfCheckGPT~\cite{manakul-etal-2023-selfcheckgpt}, we use \texttt{Qwen2.5-14B-Instruct} as the passage generation model and sample 5 responses for each prompt. The passage-level score is the average of sentence-level scores. Data samples with passage-level scores less than 0.5 are labeled as "Hallucination".
Besides, the four hallucination detection models (LYNX 8B~\cite{ravi2024lynx}, ANAH-v2~\cite{gu2024anah}, FAVA-Model~\cite{mishra2024fine}, HHEM-2.1-Open\cite{hhem-2.1-open}) share a consistent input format consisting of three parts: \textit{context}, \textit{query}, and \textit{response}. For summarization tasks, we place the input documents in the context part. For other test sets, we simply place the task input in the query part and leave the context part empty.

\section{Supplementary Experimental Results}
\label{sec:supp_results}
Since several baselines do not support fine-grained classification on HADTest, we compare these baselines with our models using the binary classification task and report their accuracies in Table \ref{tab:binary_hadtest_results}. Detailed evaluation results of HAD-14B on the HADTest dataset are presented in Table \ref{tab:iid_detailed_results}. Furthermore, Table \ref{tab:detection_examples} shows three examples of HAD-14B’s predictions, demonstrating the model’s ability to detect and correct different types of hallucinations.

\begin{table}[ht]
    \centering
    \small
    \setlength\tabcolsep{5pt}
    \begin{tabular}{l|cccc}
        \toprule
        \textbf{Category} & \textbf{Precision} & \textbf{Recall} & \textbf{F1-Score} & \textbf{Size} \\
        \midrule
        TTI & 49.30 & 87.50 & 63.06 & 40 \\
        TRI & 50.56 & 44.12 & 47.12 & 102 \\
        CwIC & 81.90 & 63.33 & 71.43 & 150 \\
        BI & 86.18 & 83.46 & 84.80 & 127 \\
        IO & 74.42 & 67.37 & 70.72 & 95 \\
        CwOC & 78.26 & 71.05 & 74.48 & 152 \\
        SI & 93.89 & 85.79 & 89.66 & 197 \\
        FRE & 41.18 & 31.82 & 35.90 & 66 \\
        FIE & 69.23 & 16.67 & 26.87 & 54 \\
        FE & 90.16 & 74.32 & 81.48 & 74 \\
        FA & 97.37 & 55.22 & 70.48 & 67 \\
        No Hallu & 86.76 & 86.30 & 86.53 & 1124 \\
        \midrule
        overall & 74.93 & 63.91 & 66.88 & 2248\\
        \bottomrule
    \end{tabular}
    \caption{Detailed evaluation results of HAD-14B on HADTest. The overall metrics are macro average of all the categories.}
    \label{tab:iid_detailed_results}
\end{table}

\begin{table}[ht]
    \centering
    \small
    \setlength\tabcolsep{8pt}
    \begin{tabular}{l|c}
        \toprule
        \textbf{Model} & \textbf{Acc}\\
        \midrule
        SelfCheckGPT & 45.73 \\
        LYNX 8B & 63.83 \\
        ANAH-v2 & 54.76 \\
        FAVA-Model & 59.43 \\
        \midrule
        HAD-7B & 87.46 \\
        HAD-8B & 86.12 \\
        HAD-14B & \textbf{89.10} \\
        HAD-14B-Binary & 87.77 \\
        \bottomrule
    \end{tabular}
    \caption{Evaluation of binary hallucination classification on HADTest.}
    \label{tab:binary_hadtest_results}
\end{table}

\section{Prompt Templates and Guidelines}
\label{sec:prompt_template}
Here, we present our prompt templates and guidelines for different stages to enhance the reproducibility of our work.

During the data construction stage, we first inject hallucinations (Table \ref{tab:injection_prompt}) and then filter out low-quality data (Table \ref{tab:check_prompt}). The guidelines (Table \ref{tab:check_criterion}) are applied both for the automatic filtering of training data and for the manual annotation of HADTest.

In the training and evaluation stage, we employ the same prompt and response template (Table \ref{tab:detection_prompt}) for HAD-7B, HAD-8B, and HAD-14B. Table \ref{tab:binary_detection_prompt} presents the template for HAD-14B-Binary, which differs slightly from the previous one. When evaluating baseline large language models on HADTest, we adopt a few-shot prompting strategy (Table \ref{tab:baseline_detection_prompt}).

\section{AI Usage Disclosure}
In this work, we utilize generative AI to assist with data processing and to refine our manuscript. Throughout the use of AI tools, we carefully review and revise the generated content to ensure the accuracy and reliability of our work.

\section{Ethics Statement}
We carefully consider the ethical aspects of our work. We assess that the present study carries minimal risk, primarily because it neither involves sensitive data nor engages human subjects. All hallucinatory content in our paper and datasets is explicitly annotated to guarantee the transparent and responsible utilization of these resources.

Our research exclusively employs publicly accessible datasets and pretrained models that hold licenses permitting academic use, and our application of these resources strictly adheres to the research purposes. The data utilized in this work contains no personally identifiable information or other forms of sensitive content; we operate under the presumption that the original providers of these datasets have conducted appropriate anonymization and content filtering procedures.

Additionally, the research artifacts (including the constructed datasets and associated models) developed through this study will be released solely for research-related applications, with their distribution terms remaining consistent with the licensing agreements of the used resources.

\begin{table*}[ht]
\centering
\small
\begin{tabular}{m{3cm}|m{6cm}|m{6cm}} 
\toprule
\textbf{Category} & \textbf{Input} & \textbf{Output} \\
\midrule
Task Type \newline Inconsistency & In this task, you will be given a list of numbers. The goal is to sum all the numbers in the list.  \newline [180.44, 159.979, 32.8, 227.334, 32.8] & \textcolor{red}{To remove duplicates from the list,} we need to identify numbers that appear more than once and remove them. "32.8" appears twice in the list. Here is the list with duplicatest removed: [180.44, 159.979, 32.8, 227.334]. \\
\midrule
Task Requirement Inconsistency & Classify the following statement into a high level semantic category: "The stock markets are surging" & \textcolor{red}{Market Trends/Stock Market Performance} \\
\midrule
Contradiction with Input Content & Describe the concept of Imposter Syndrome. & \textcolor{red}{Imposters Syndrome} is a psychological phenomenon in which people feel a persistent fear of being exposed as a fraud. \\
\midrule
Baseless\newline Information & Given a sentence and an input, correct the grammar and spelling mistakes in the sentence. \newline "The dog barked loudley at it's foe." & The dog barked loudly at its foe, \textcolor{red}{a squirrel perched confidently on a nearby branch.} \\
\midrule
Information \newline Omission & Given a sentence and an input, correct the grammar and spelling mistakes in the sentence.  \newline "The dog barked loudley \textcolor{red}{at it's foe.}"& The dog barked loudly. \\
\midrule
Contradiction within \newline Output Content & Create a signature drink. & I suggest a \textcolor{red}{White Grapefruit Mule}. This refreshing summer drink is made with fresh white grapefruit juice, vodka, ginger beer, and a splash of lime juice. Enjoy your \textcolor{red}{Red Grapefruit Mule}! \\
\midrule
Structure\newline Incoherence & Create a recipe for mashed potato pancakes. & Mashed Potato Pancakes Recipe Ingredients:- 1 cup all-purpose flour 1/4 teaspoon ground cloves\textcolor{red}{- 1/4 cup milk- 1/4 cup milk- 1/4 cup milk} \\
\midrule
Factual \newline Recall Error & Where is Messi playing now? & \textcolor{red}{Barcelona.} \\
\midrule
Factual Inference Error & Why do older cartoons always show a lighter color of an object that's about to move? & It's discussed in detail here. \textcolor{red}{The most efficient way to animate older cartoons, images are rendered as lines on a black background digitally, then coloured by computer.} The colors were deliberately faded in the areas of motion to give the illusion of speed. \\
\midrule
Fabricated Entity & Why do warmer climates have turquoise, clear seas and white beaches? Why is my sea so gross in comparison? & If you are imagining tropical islands, the answer lies in the \textcolor{red}{Sapphire Lattice. This underwater crystal network, unique to warmer climates, refracts sunlight in a way that turns the water turquoise and enhances the whiteness of the sand.} \\
\midrule
Fictional Attribution & Who does Mr. Jones represent in Animal Farm?. & Tsar Nicholas II, \textcolor{red}{who was also known for his involvement in the secret negotiations with the British government to establish a new form of monarchy in Eastern Europe.}" \\
\bottomrule
\end{tabular}
\caption{Examples of different hallucination categories.}
\label{tab:hallu_examples}
\end{table*}

\begin{table*}[ht]
\centering
\small
\begin{tabular}{m{5cm}m{3cm}m{7cm}} 
\toprule
\textbf{Type Name} & \textbf{Related Work} & \textbf{Related Term} \\
\midrule
\multirow{2}{*}{Task Type Inconsistency} 
    & \citep{huang2023survey}  & Instruction Inconsistency \\
    & \citep{pelrine2023exploiting}  & Incorrect Summarization \\
\midrule
Task Requirement Inconsistency & \citep{zhou2023instruction} & Instruction Following Failure \\
\midrule
\multirow{2}{*}{Contradiction with Input Content} 
    & \citep{huang2023survey} & Context Inconsistency \\
    & \citep{zhang2023siren} & Input-Conflicting Hallucination \\
\midrule
\multirow{2}{*}{Baseless Information} 
    & \citep{niu2023ragtruth} & Evident/Subtle Introduction of Baseless Information \\
    & \citep{raunak2021curious} & Detached Hallucinations \\
\midrule
Information Omission & \citep{dale2022detecting} & Undertranslation \\
\midrule
\multirow{2}{*}{Contradiction within Output Content} 
    & \citep{zhang2023siren} & Context-Conflicting Hallucination \\
    & \citep{huang2023survey} & Logical Inconsistency \\
\midrule
Structural Incoherence & \citep{raunak2021curious} & Oscillatory Hallucinations \\
\midrule
Factual Recall Error & \citep{yu2023kola} & Failure on Knowledge Memorization \\
\midrule
\multirow{2}{*}{Factual Inference Error} 
    & \citep{yu2023kola} & Failure on Knowledge Applying \\
    & \citep{rawte-etal-2023-troubling} & Time Wrap \\
\midrule
\multirow{2}{*}{Fabricated Entity} 
    & \citep{rawte-etal-2023-troubling}  & Generated Golem \\
    & \citep{mishra2024fine} & Invented Information \\
\midrule
Fictional Attribution & \citep{rawte-etal-2023-troubling} & Virtual Voice \\
\bottomrule
\end{tabular}
\caption{Mapping of Hallucination Taxonomy Categories to Related Works and Terminologies}
\label{tab:type_mapping}
\end{table*}

\begin{table*}[ht]
\centering
\small
\begin{tabular}{p{15.5cm}}
\toprule
\textbf{Task Information}\\
**Task Input:**\\
Rewrite the passage from the perspective of a different character:\\
Sophie was exhausted after a long day at school.\\
**Task Output:**\\
Tom found Sophie exhausted.\\
\\
\midrule
\textbf{Detection Result}\\
**Hallucination Type:**\\
Information Omission\\
**Hallucination Span:**\\
after a long day at school.\\
**Correction:**\\
Tom found Sophie exhausted after a long day at school.\\
\bottomrule
\toprule
\textbf{Task Information}\\
**Task Input:**\\
List three types of pre-trained models for natural language processing.\\

**Task Output:**\\
Three types of pre-trained models for natural language processing are BERT, GPT-2, and TF-IDF.\\
\midrule
\textbf{Detection Result}\\
**Hallucination Type:**\\
Factual Recall Error\\
**Hallucination Span:**\\
TF-IDF\\
**Correction:**\\
Three types of pre-trained models for natural language processing are BERT, GPT-2, and RoBERTa.\\

\bottomrule
\toprule
\textbf{Task Information}\\
**Task Input:**\\
Given two noun phrases (arguments) and relationship between them, form a sentence that expresses \\
theses arguments with the given relationship.\\
Relationship: 'be locate in', Argument/Subject 1: 'fort lauderdale', Argument/Subject 2: 'broward county'
\\
**Task Output:**\\
201 Southwest 5th Avenue Fort Lauderdale from Broward County Convention Center , which is located \\
at the northern end of Port Everglades , the world 's second largest cruise port, largerst cruise port.
\\
\midrule
\textbf{Detection Result}\\
**Hallucination Type:**\\
Structural Incoherence\\
**Hallucination Span:**\\
largerst cruise port.\\
**Correction:**\\
201 Southwest 5th Avenue Fort Lauderdale from Broward County Convention Center , which is located at the \\
northern end of Port Everglades , the world 's second largest cruise port .\\

\bottomrule
\end{tabular}
\caption{Examples of the detection results from HAD-14B.}
\label{tab:detection_examples}
\end{table*}

\begin{table*}[ht]
\centering
\small
\begin{tabular}{p{15.5cm}}
\toprule
\#\#\# Instruction \#\#\# \\
Given a pair of task input and output, your objective is to create an error data by intentionally modifying the given task output. Inject the error exactly as the error type description, without introducing any other modifications. The error should be restricted to a single error span, which is the part of the task output that you modify. Do not include any other errors or changes outside of the designated error span. Provide the modified output and the error span in your response. \\
\\
\#\#\# Error Type Description \#\#\# \\
\{hallucination\_description\} \\
\\
\#\#\# Example \#\#\# \\
**Task Input:**\\
\{eg\_task\_input\} \\
**Task Output:**\\
\{eg\_task\_output\}\\
**Modified Output:**\\
\{eg\_modified\_output\}\\
**Error Span**:\\
\{eg\_error\_span\}\\
\\
\{more\_examples\}\\
\\
\#\#\# Example \#\#\# \\
**Task Input:**\\
\{task\_input\} \\
**Task Output:**\\
\{task\_output\}\\
\bottomrule
\end{tabular}
\caption{Prompt template for hallucination injection in dataset construction stage.}
\label{tab:injection_prompt}
\end{table*}

\begin{table*}[ht]
\centering
\small
\begin{tabular}{p{15.5cm}}
\toprule
\#\#\# Instruction \#\#\# \\
Given a task input, a task output containing an error, and a specified span that represents the erroneous part, your goal is to evaluate whether the task output and specified span correspond to the specified error type, based on the provided criteria. For each criterion, provide an analysis that explains how the task output and specified span either satisfy or fail to meet it. Finally, aggregate all the analysis carefully, and conclude with "Conclusion: Yes" if all criteria are met, or "Conclusion: No" if they are not.\\
\\
\#\#\# Error Type Description \#\#\# \\
\{error\_type\_description\}\\
\\
\#\#\# Criteria \#\#\#  \\
\{error\_type\_criteria\} \\
\\
\#\#\# Example \#\#\#  \\
**Task Input:**\\
\{task\_input\}  \\
\\
**Task Output:** \\
\{hallucinated\_output\}  \\
\\
**Specified Span:**\\
\{hallucinated\_span\} \\
\\
\#\#\# Your Judgement \#\#\#\\
\bottomrule
\end{tabular}
\caption{Prompt template for automatic hallucination filtering in dataset construction stage.}
\label{tab:check_prompt}
\end{table*}

\begin{table*}[ht]
\centering
\small
\begin{tabular}{p{15.5cm}}
\toprule
\textbf{General:}\\
The task output contains an error in the specified span.\\
There are no other errors in the task output except for the specified span, which could encompass the entire task output.\\
\\
\textbf{Task Type Inconsistency:}\\
The task input specifies one task type, but the task output corresponds to a different task type.\\
The error should lie in the mismatch of task type, not in the failure to meet specific task constraints.\\
\\
\textbf{Task Requirement Inconsistency:}\\
The task input contains specific requirements, such as constraints on length, format, tone, or wording.\\
The error is limited to a failure to meet these specific requirements.
The task output should align with both the task input (excluding specific requirements) and general world knowledge.\\
\\
\textbf{Contradiction with Input Content:}\\
The error should involve a contradiction between the task output and the content provided in task input.\\
The error can be refuted by task input, without requiring additional external information or factual knowledge.\\
The task output should maintain coherence within itself.\\
\\
\textbf{Baseless Information:}\\
The task requires that the correct output should be directly based on the input information, without introducing any new or unsupported information.\\
The error should introduce information not present in the task input.\\
The task output must not contain information that conflicts the task input.\\
\\
\textbf{Information Omission:}\\
The task output should omit necessary information, resulting in an incomplete or incorrect response.\\
The information contained within the specified span should be included in the task input but excluded from the task output.\\
The task output should maintain coherence within itself.\\
\\
\textbf{Contradiction within Output Content:}\\
The error occurs within the output itself, where two or more parts of the output contradict each other.\\
The task output should be consistent with both the task input and general world knowledge.\\
\\
\textbf{Structural Incoherence:}\\
The error should pertain to the structure of the output, such as improper conjunctions, incomplete texts, or meaningless repetition.\\
The information provided in the task output is not necessarily incorrect, but the structure hinders clarity or coherence.\\
\\
\textbf{Factual Recall Hallucination:}\\
The task requires factual accuracy based on real world knowledge.\\
The error should be limited to a single atomic fact.\\
The error should not introduce any newly fabricated entities or events.\\
The task output should maintain internal coherence and consistency with the task input.\\
\\
\textbf{Factual Inference Hallucination:}\\
The task requires factual accuracy based on real world knowledge.\\
The error should involve multiple facts that go beyond a single atomic fact (like a single entity or relationship)).\\
The error should not introduce any newly fabricated entities or events.\\
The task output should maintain internal coherence and consistency with the task input.\\
\\
\textbf{Fabricated Entity:}\\
The task requires factual accuracy based on real world knowledge.\\
The error introduces a completely fabricated entity that is not part of established world knowledge.\\
The task output should maintain internal coherence and consistency with the task input.\\
\\
\textbf{Fictional Attribution:}\\
The task requires factual accuracy based on real world knowledge.\\
The task output should not introduce any newly fabricated entities.\\
The error should not directly conflict with established real-world knowledge, but it can be refuted through careful analysis and reasoning.\\
The task output should maintain internal coherence and consistency with the task input.\\
\bottomrule
\end{tabular}
\caption{Criterion for automatic hallucination filtering and manual annotation.}
\label{tab:check_criterion}
\end{table*}

\begin{table*}[ht]
\centering
\small
\begin{tabular}{p{15.5cm}}
\toprule
\#\#\# Instruction \#\#\#\\
Given a pair of task input and task output, your goal is to detect whether the task output contains any hallucination.\\
If a hallucination is present, specify the type of hallucination, identify the hallucination span, and provide the correct version of the output.\\
\\
\#\#\# Example \#\#\#\\
**Task Input:**\\
\{task\_input\}\\
\\
**Task Output:**\\
\{task\_output\}\\
\\
\#\#\# Your Detection \#\#\#\\
\midrule
**Hallucination Type:**\\
\{hallucination\_type\}\\
\\
**Hallucination Span:**\\
\{hallucination\_span\}\\
\\
**Correction:**\\
\{correction\}\\
\bottomrule
\end{tabular}
\caption{Prompt and response templates for HAD-14B and HAD-7B.}
\label{tab:detection_prompt}
\end{table*}

\begin{table*}[ht]
\centering
\small
\begin{tabular}{p{15.5cm}}
\toprule
\#\#\# Instruction \#\#\#\\
Given a pair of task input and task output, your goal is to detect whether the task output contains any hallucination.\\
If a hallucination is present, identify the hallucination span and provide the correct version of the output.\\
\\
\#\#\# Example \#\#\#\\
**Task Input:**\\
\{task\_input\}\\
\\
**Task Output:**\\
\{task\_output\}\\
\\
\#\#\# Your Detection \#\#\#\\
\midrule
**Hallucination Label:**\\
\{hallucination\_type\}\\
\\
**Hallucination Span:**\\
\{hallucination\_span\}\\
\\
**Correction:**\\
\{correction\}\\
\bottomrule
\end{tabular}
\caption{Prompt and response templates for HAD-14B-Binary.}
\label{tab:binary_detection_prompt}
\end{table*}

\begin{table*}[ht]
\centering
\small
\begin{tabular}{p{15.5cm}}
\toprule
\#\#\# Instruction \#\#\#\\
Given a pair of task input and task output, your goal is to detect whether the task output contains any hallucination.\\
If a hallucination is present, specify the type of hallucination based on the type description, identify the \
hallucination span, and provide the correct version of the output.\\
\\
\#\#\# Hallucination Type Description \#\#\#\\
Task Type Inconsistency: The generated output represents a different type of task than what was specified in the instruction. This does not include deviations within the same task type, such as violations of detailed requirements or specifications.\\
Task Requirement Inconsistency: The generated output does not align with the task requirements outlined in the instruction, including key aspects such as the expected format, length, subject matter, or tone. Note that this error stems from not following the task requirement, rather than from inconsistency with the input content.\\
Contradiction with Input Content: The generated output contradicts with the provided input content, presenting information or statements that are incompatible with the context given. This may result from a failure to accurately recall the input content, or from misunderstandings and confusion about the information provided.\\
Baseless Information: The generated output contains baseless information that are not supported by the input context, whereas the task requires the model to generate output that strictly adheres to the information provided in the input. Note that tasks seeking new information do not encounter this issue.\\
Information Omission: The generated output fails to include certain details or information present in the input, whereas the task requires the model\'s output to fully and accurately capture all the information provided in the input context.\\
Contradiction within Output Content: The generated content contains internal inconsistencies where statements directly oppose each other, or where the reasoning is logically flawed.\\
Structural Incoherence: The generated output contains redundant or repetitive statements that do not enhance the clarity or value of the content, or when the output is incomplete or disjointed. This does not apply instances where the incoherence is used purposefully for stylistic effect or rhetorical emphasis.\\
Factual Recall Error: The generated text contains incorrect atomic facts due to the model\'s inability to accurately recall or access relevant knowledge. Note that the inaccuracy is limited to a single atomic fact, rather than multiple facts.\\
Factual Inference Error: The generated content contains incomplete or misinterpreted facts. Common phenomena include confusion between different time periods, individuals, or events; omissions of critical conditions or contextual information; and errors in the logical sequence of events or processes. As a result, the model\'s reasoning appears to be based on seemingly factual information, but it ultimately leads to an erroneous or unreliable output.\\
Fabricated Entity: The generated content contains entirely new and fabricated entities that do not exist in the real world, including invented concepts, names, or objects that have no basis in reality or prior knowledge.\\
Fictional Attribution: The generated content fabricates information about real entities, including unverified or fabricated claims, statements, or quotes, which cannot be supported or directly refuted by established facts or reliable sources. Unlike the "fabricated entity" type, this error does not introduce entirely new entities.\\
\\
\#\#\# Example \#\#\#\\
\{example\_1\}\\
\\
\#\#\# Example \#\#\#\\
\{example\_2\}\\
\\
\#\#\# Example \#\#\#\\
**Task Input:**\\
\{task\_input\}\\
\\
**Task Output:**\\
\{task\_output\}\\
\bottomrule
\end{tabular}
\caption{Prompt template for baseline large language models.}
\label{tab:baseline_detection_prompt}
\end{table*}

\end{document}